\pdfoutput=1

\documentclass[11pt]{article}

\usepackage{acl}

\usepackage{times}
\usepackage{latexsym}

\usepackage[T1]{fontenc}

\usepackage[utf8]{inputenc}

\usepackage{microtype}

\usepackage{inconsolata}

\usepackage{graphicx}
\usepackage{svg}
\usepackage{algorithm}
\usepackage{algpseudocode}
\usepackage{amsmath}
\usepackage{booktabs}
\usepackage{multirow}
\usepackage{enumitem}
\usepackage{xspace}

\usepackage{soul}  

\definecolor{MyRed}{HTML}{D63F3F}
\definecolor{MyBlue}{HTML}{265ED4}
\definecolor{MyGreen}{HTML}{0A8F6B}
\definecolor{MyPurple}{HTML}{9467bd}
\definecolor{MyOrange}{HTML}{FF9100}
\definecolor{MyGrey}{HTML}{616C7A}

\title{TOPICAL: TOPIC Pages AutomagicaLly}

\author{
    John Giorgi$^{1,2,3}$\thanks{\enspace Work performed during internship at AI2} \quad Amanpreet Singh$^{4}$ \quad \textbf{Doug Downey}$^{4, 5}$ \\\vspace{4pt} 
    \textbf{Sergey Feldman}$^{4}$ \quad \textbf{Lucy Lu Wang}$^{4,6}$\\\vspace{4pt}
  $^{1}$University of Toronto \quad
  $^{2}$Terrence Donnelly Centre \quad
  $^{3}$Vector Institute for AI \\
  $^{4}$Allen Institute for AI \quad
  $^{5}$Northwestern University \quad $^{6}$University of Washington \\
  {\footnotesize \bf\texttt{john.giorgi@utoronto.ca, lucylw@uw.edu, \{sergey, dougd\}@allenai.org}}
}

\begin{document}

\maketitle

\begin{abstract}
Topic pages aggregate useful information about an entity or concept into a single succinct and accessible article. Automated creation of topic pages would enable their rapid curation as information resources, providing an alternative to traditional web search. While most prior work has focused on generating topic pages about biographical entities, in this work, we develop a completely automated process to generate high-quality topic pages for scientific entities, with a focus on biomedical concepts. We release TOPICAL, a web app and associated open-source code, comprising a model pipeline combining retrieval, clustering, and prompting, that makes it easy for anyone to generate topic pages for a wide variety of biomedical entities on demand. In a human evaluation of 150 diverse topic pages generated using TOPICAL, we find that the vast majority were considered relevant, accurate, and coherent, with correct supporting citations. We make all code publicly available and host a free-to-use web app at: \url{https://s2-topical.apps.allenai.org}.
\end{abstract}

\section{Introduction}

\begin{figure}[t]
    \includegraphics[width=\columnwidth]{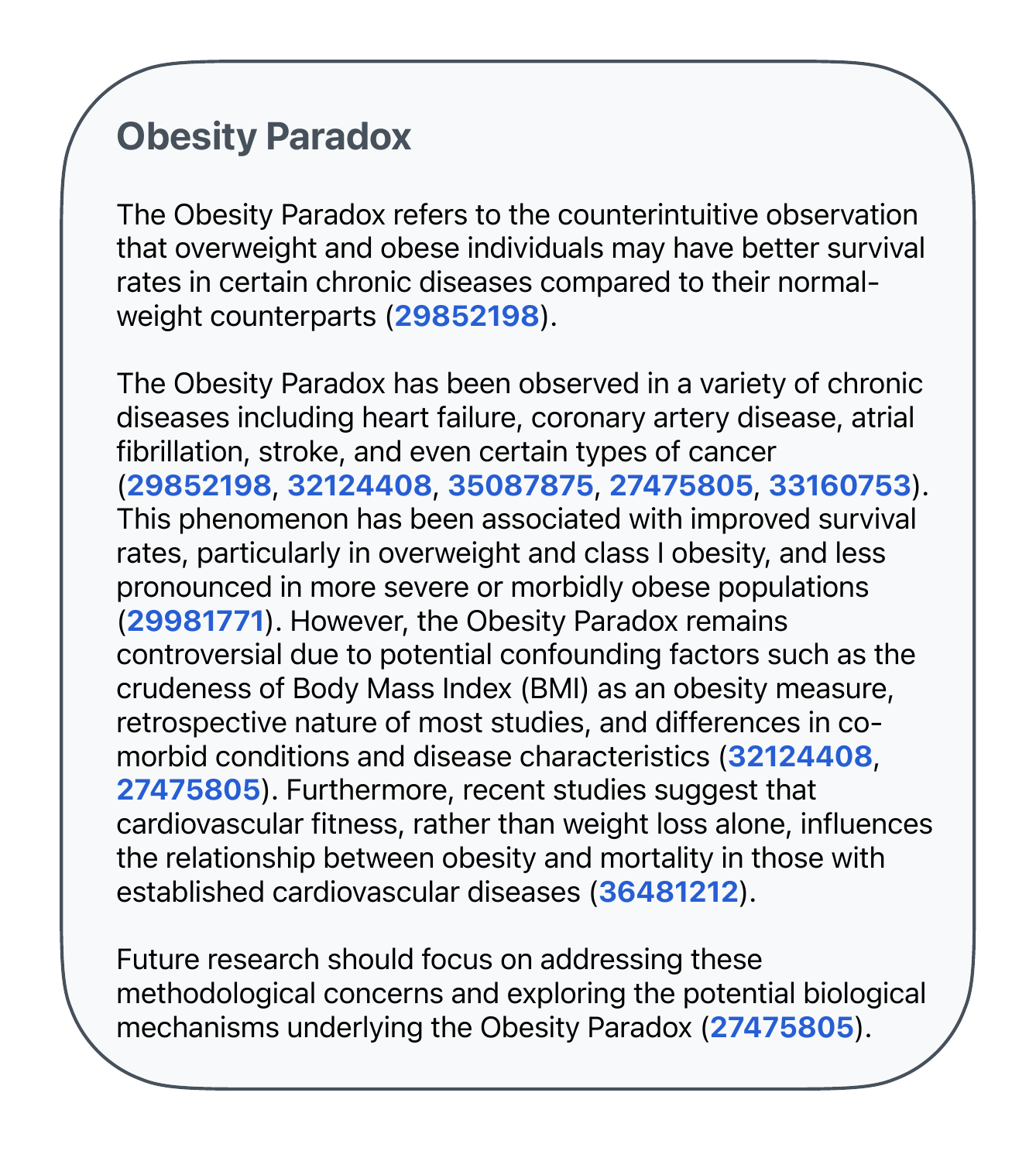}
    \caption{Example of a scientific topic page generated by our system. Citations are provided as hyperlinks to PubMed articles and denoted by their PMID. The topic page is divided into the definition statement, main content, and future directions and open research questions.}
    \label{fig:topic-page}
    \vspace{-2mm}
\end{figure}

The automatic generation of topic pages is a long-standing goal of the NLP community \cite{10.1145/1645953.1646298, Balasubramanian2010BeyondRL, 5629127, Pochampally_Karlapalem_Yarrabelly_2021}. In contrast to web search results---displayed as ranked lists of hyperlinks with short text snippets across many pages---topic pages aggregate useful information about various aspects of an entity or concept in a single, concise location. \textit{Scientific} topic pages \cite{Wodak2012TopicPP, 10.1007/978-3-031-28238-6_23} apply this thinking to scientific concepts by aggregating information from the primary literature to produce succinct and accessible summaries useful to both experts and non-experts alike (\autoref{fig:topic-page}). Among other things, high-quality scientific topic pages hold the promise of:

\begin{enumerate}[itemsep=0.2pt, topsep=3pt, leftmargin=10pt]
    \item \textbf{Helping manage the torrent of scientific literature}. 
A staggering amount of scientific information is published daily. In biomedicine alone, nearly 4,000 papers (>2 per minute) are deposited in PubMed or bioRxiv each day, leading to a general state of ``information overload'' \cite{Landhuis2016ScientificLI, hope2023computational}. Automatically generated topic pages allow researchers to quickly familiarize themselves with an area and its most active research directions, while citations to source articles provide an entry-point into the literature for in-depth exploration.\footnote{Topic pages generated by our system provide citations to highly relevant primary literature. See \textsection \ref{clustering} for details.}
    \item \textbf{Improving the accessibility of scientific texts}. Encyclopedic resources like Wikipedia contain descriptions for a small fraction of scientific concepts \cite{King2020HighPrecisionEO}. Therefore, non-expert readers may turn to the primary literature for information \cite{August2022PaperPM}, e.g., a patient or caregiver wishing to learn about a new drug or rare disease. However, most scientific text assumes extensive background knowledge that a non-expert reader is unlikely to possess \cite{Portenoy2021BurstingSF, murthy-etal-2022-accord}. Automatically generated topic pages hold the promise of improving the accessibility of scientific texts, both by providing an \textit{alternative} to the primary literature and by serving as a \textit{resource} to help fill in the gaps in a reader's background knowledge.
\end{enumerate}

\noindent In this work, we develop a fully automated process leveraging large language models (LLMs) to generate high-quality scientific topic pages, with a focus on biomedical topics (\textsection \ref{approach}). Our solution is available as an easy-to-use and publicly available web app (\textsection \ref{web-app}), and associated source code.\footnote{\url{https://github.com/allenai/TOPICAL}} We validate the quality of TOPICAL via extensive human evaluation on 150 diverse biomedical terms from the MeSH\footnote{Medical Subject Headings (MeSH) is a hierarchical vocabulary used to index articles and books in the life sciences.} hierarchy (\textsection \ref{human-eval}) and find that the vast majority of topic pages are rated as relevant, accurate, and coherent, with correct citations to primary sources (\textsection \ref{results}).

\section{Related Work}

\paragraph{Topic page generation} Topic page generation is usually framed as a topic-focused, open-domain multi-document summarization (MDS) task \cite{giorgi-etal-2023-open}. Most prior work is concerned with generating Wikipedia-like pages for general-domain entities and concepts (often biographical in nature). Early work clustered the web search query logs for an entity of interest to determine its various aspects, used each aspect cluster to retrieve and rank relevant sentences, and then re-organized the retrieved sentences for coherence to produce a bullet-list style topic page \cite{10.1145/1645953.1646298, Balasubramanian2010BeyondRL, 5629127}.

More recent work---also focused on biographical entities---first templates the topic page by copying common section headings from Wikipedia pages for related topics and trains a supervised model to select the text content for each section. An unsupervised component then creates topic-specific sections, and several post-processing steps are applied to reduce redundancy and improve coherence \cite{Pochampally_Karlapalem_Yarrabelly_2021}. In contrast, our work focuses on topics of scientific interest, does not try to match a Wikipedia-like structure, and generates topic pages in a \textit{abstractive} fashion.

\paragraph{Scientific topic pages}
\citet{10.1007/978-3-031-28238-6_23} investigate generating scientific topic pages at scale; however, they do not synthesize a summary but focus rather on extracting a definition statement verbatim, alongside ``mention snippets'' and related concepts. In contrast, we attempt to synthesize more comprehensive topic pages, including a definition statement and content about the entity's main and future research directions. \citet{king-etal-2022-dont} introduce a Scientific Concept Description task with similar motivation to our work, but focus on earlier, smaller generative models for describing computer science concepts, and find the systems to hallucinate relatively frequently. WikiCrow, based on PaperQA \citep{Lala2023PaperQARG}, provides scientific topic pages generated by an LLM-based system for human protein-coding genes. In contrast to our approach, their publicly available demo is limited to 15,616 pre-generated topic pages and does not allow a user to generate topic pages for a new entity of interest on demand.\footnote{https://www.futurehouse.org/wikicrow}

\begin{figure}[t]
\includegraphics[width=\columnwidth]{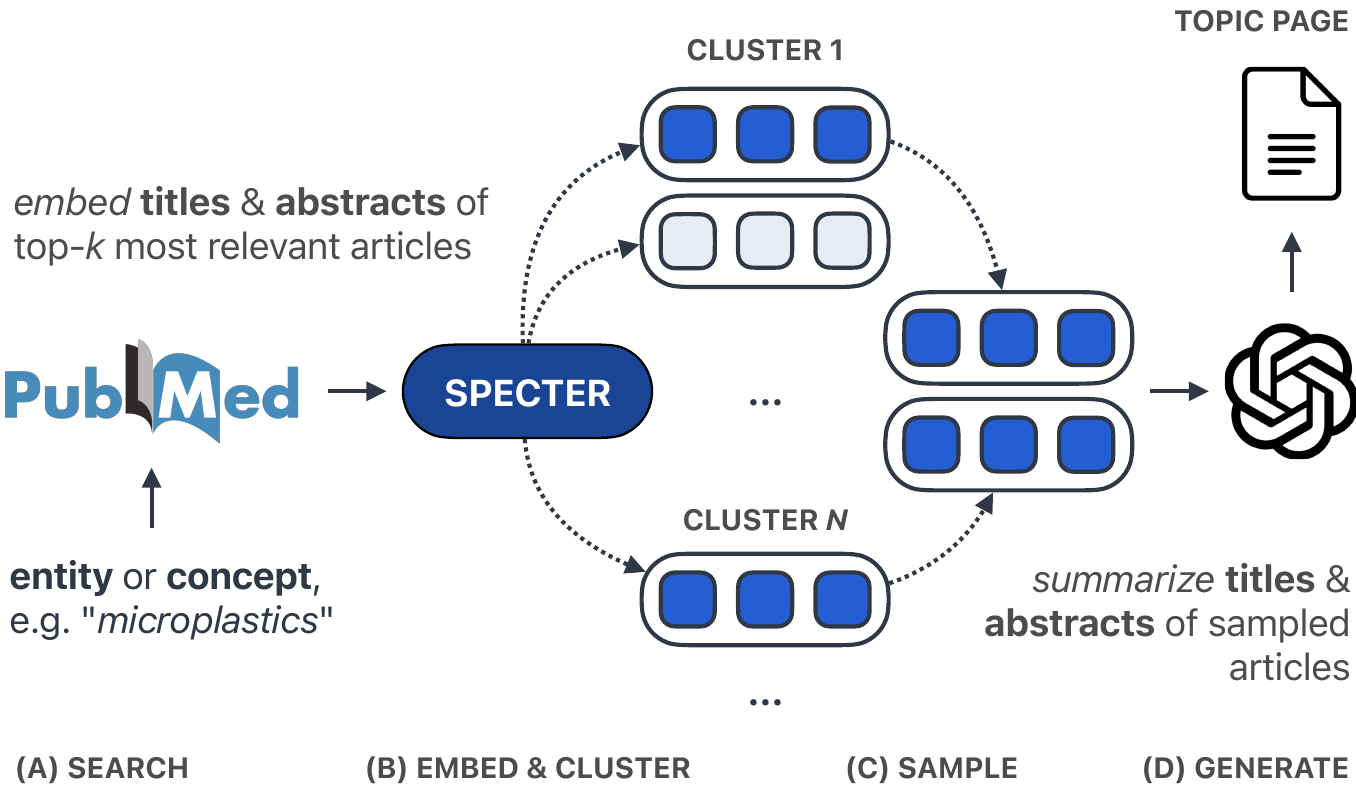}\caption{Overview of TOPICAL. Given a biomedical entity, we query PubMed for relevant literature (\textbf{A}). The titles and abstracts of the results are embedded with SPECTER \citep{singh-etal-2023-scirepeval} and clustered based on semantic similarity (\textbf{B}). We sample titles and abstracts from the clusters (\textbf{C}) and feed them to GPT-4 \citep{OpenAI2023GPT4TR}, alongside publication metadata and natural language instructions, to generate the topic page (\textbf{D}).}
\label{fig:overview}
\vspace{-1mm}
\end{figure}

\begin{figure*}[t]
\includegraphics[width=\textwidth]{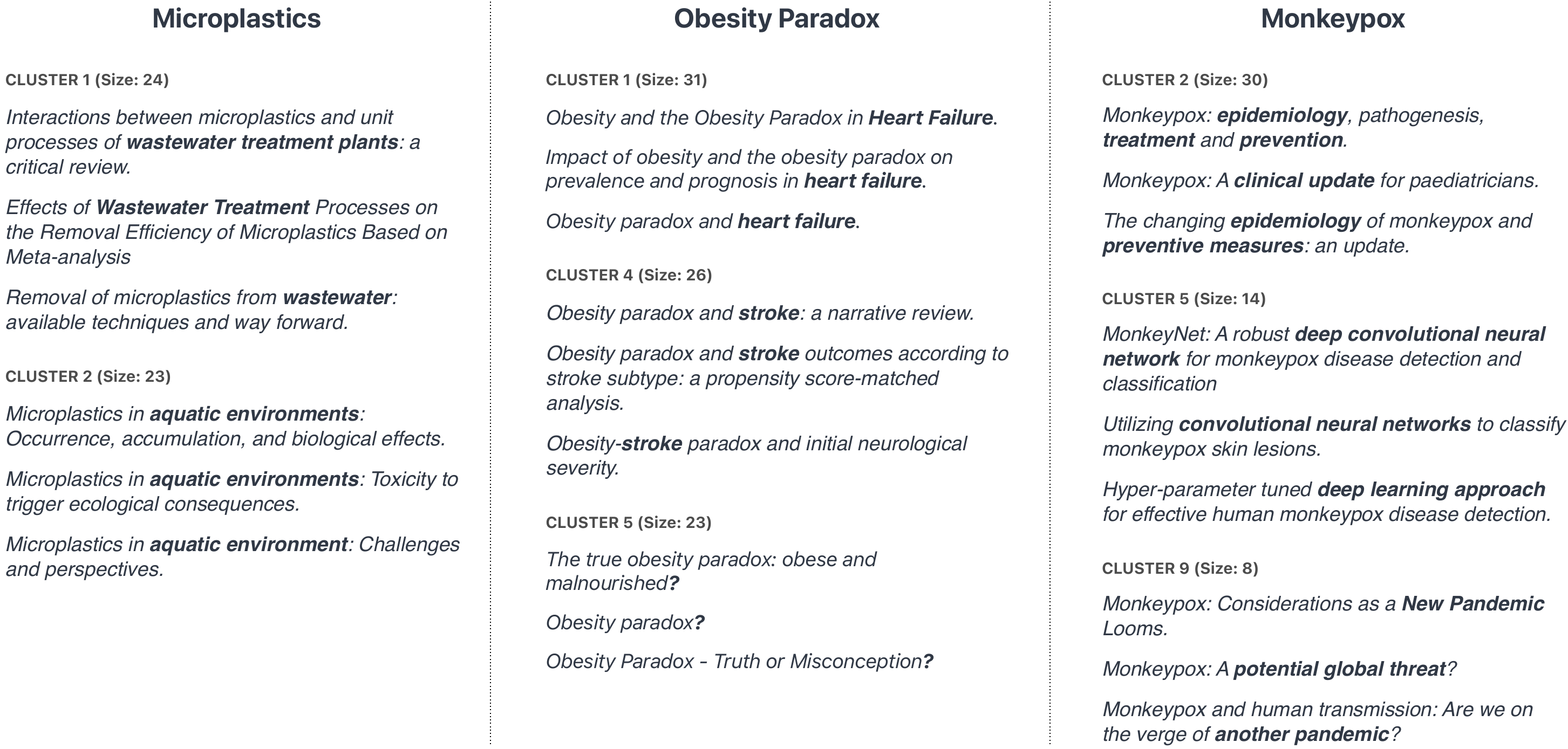}
\caption{Example clusters. Three titles from a selection of clusters for each concept are shown. \textbf{Emphasis} ours.}
\label{fig:clusters}
\end{figure*}

\section{Approach} \label{approach}

Our approach follows a retrieval-augmented generation (RAG) setup \cite{Guu2020REALMRL, Lewis2020RetrievalAugmentedGF, petroni-etal-2021-kilt, Izacard2022FewshotLW}. A large body of literature (up to 10k papers) is retrieved for a given entity (\textsection\ref{search}) and fed to a LLM alongside publication metadata and instructions (\textsection \ref{generate}). Because the amount of retrieved literature is often many times larger than the LLM's maximum context size, we design a clustering step to loosely group the literature into areas of study and sample from these clusters for input (\textsection \ref{clustering}). During prompting, the model is instructed to provide in-line citations for all claims by outputting one or more PubMed IDs (PMIDs). See \autoref{fig:overview} for an overview.

\subsection{Querying PubMed} \label{search}

The generation of each topic page begins with a user-provided biomedical entity or concept. This entity is expected to be covered by papers indexed in PubMed,\footnote{\url{https://pubmed.ncbi.nlm.nih.gov/}} a free search engine that indexes over 36 million papers on life science and biomedical topics. TOPICAL, our system, leverages the Entrez ESearch API \citep{kans2023entrez} to query PubMed and supports the full syntax of the PubMed Advanced Search Builder; however, simply inputting the entity or concept verbatim is often sufficient, e.g. ``\textit{microplastic},'' as the ESearch API will apply `automatic term mapping' (ATM)\footnote{ \url{https://pubmed.ncbi.nlm.nih.gov/help/\#automatic-term-mapping}} to this query to include, among other things, matching MeSH descriptors and pluralization (e.g. ``\textit{microplastic\textbf{s}}''). We then download the titles and abstracts of the top 10,000 most relevant papers returned by ESearch.

\subsection{Clustering and sampling the literature} \label{clustering}

The amount of retrieved literature is usually many times the maximum context size of the LLM. Therefore, we first cluster titles \& abstracts by semantic similarity to identify major areas of study, then sample from these clusters to produce a diverse set of inputs. The steps are described below:

\paragraph{Embedding} Titles and abstracts are jointly embedded using the SPECTER2 PRX model  \citep{singh-etal-2023-scirepeval}, a text encoder specifically designed for producing highly-quality representations of scientific text from a paper's title and abstract. We formatted each input as: ``\texttt{\{title\}} [SEP] \texttt{\{abstract\}}''.

\paragraph{Clustering} We apply a clustering algorithm which identifies `communities': clusters of embeddings of a minimum size with a pairwise cosine similarity greater than or equal to some threshold.\footnote{\url{https://www.sbert.net/examples/applications/clustering/README.html\#fast-clustering}} We set the similarity threshold to \texttt{0.96} and the minimum cluster size to \texttt{5}. In degenerate cases where fewer than 2 clusters are identified, we iteratively reduce the similarity threshold by \texttt{0.02}, stopping when at least 2 clusters are identified or the threshold falls below \texttt{0.90}---in which case we skip the clustering step. See \autoref{fig:clusters} for examples of clusters produced by this process.

\begin{algorithm}
\caption{Sampling Procedure for Papers}
\label{alg:sampling}
\small
\begin{algorithmic}[1]
\Require Collection of clustered titles + abstracts, $\mathcal{C}$
\Require Maximum number of input tokens, $T_{\text{max}}$

\State $\mathcal{C} \gets \text{sorted}(\mathcal{C})$ \Comment{By descending cluster size}
\State Initialize $\mathcal{S} \gets \emptyset$ \Comment{Sampled papers}
\State $t \gets 0$  \Comment{Current token count}

\For{each $C_i$ in $\mathcal{C}$}
    \State $c \gets \text{centroid of } C_i$
    \If{$t + |c| \leq T_{\text{max}}$}
        \State Append $c$ to $\mathcal{S}$
        \State $t \gets t + |c|$  \Comment{$|c|$ is the number of tokens in $c$}
    \EndIf
\EndFor

\While{$t < T_{\text{max}}$ and there exist unsampled papers in $\mathcal{C}$}
    \State Sample a paper $p$ from $\mathcal{C}$ with a probability $\propto \sqrt{|C_i|}$
    \If{$t + |p| \leq T_{\text{max}}$}
        \State Append $p$ to $\mathcal{S}$
        \State $t \gets t + |p|$  \Comment{$|p|$ is the number of tokens in $p$}
    \EndIf
\EndWhile

\Ensure Return $\mathcal{S}$ as a list of lists (outer: unique clusters, inner: papers from the cluster)
\end{algorithmic}
\end{algorithm}

\paragraph{Sampling} We sample as many titles and abstracts as will fit in the prompt to the LLM. If the number of papers returned in the search step is 100 or less, or no clusters were identified in the clustering step, we randomly sample papers for inclusion. Otherwise, we do the following: first, sort clusters by decreasing size. Then, select each \textit{centroid} for inclusion, starting with the largest cluster and continuing until the centroids of all clusters have been selected or the model's maximum input size has been reached. If all centroids have been selected and the model's maximum input tokens are not exhausted, we sample from the remaining clusters with a probability proportional to the square root of the cluster size (see Algorithm~\ref{alg:sampling} for details).\footnote{\url{https://en.wikipedia.org/wiki/Square\_root\_biased_sampling}} This sampling strategy is motivated by the idea that we should aim to capture as many and as diverse areas of study for a concept as possible (hence the selection of centroids) while favouring more commonly studied subtopics (hence the weighted sampling).

\subsection{Generating the topic page} \label{generate}

\begin{figure}[th!]
\includegraphics[width=\columnwidth]{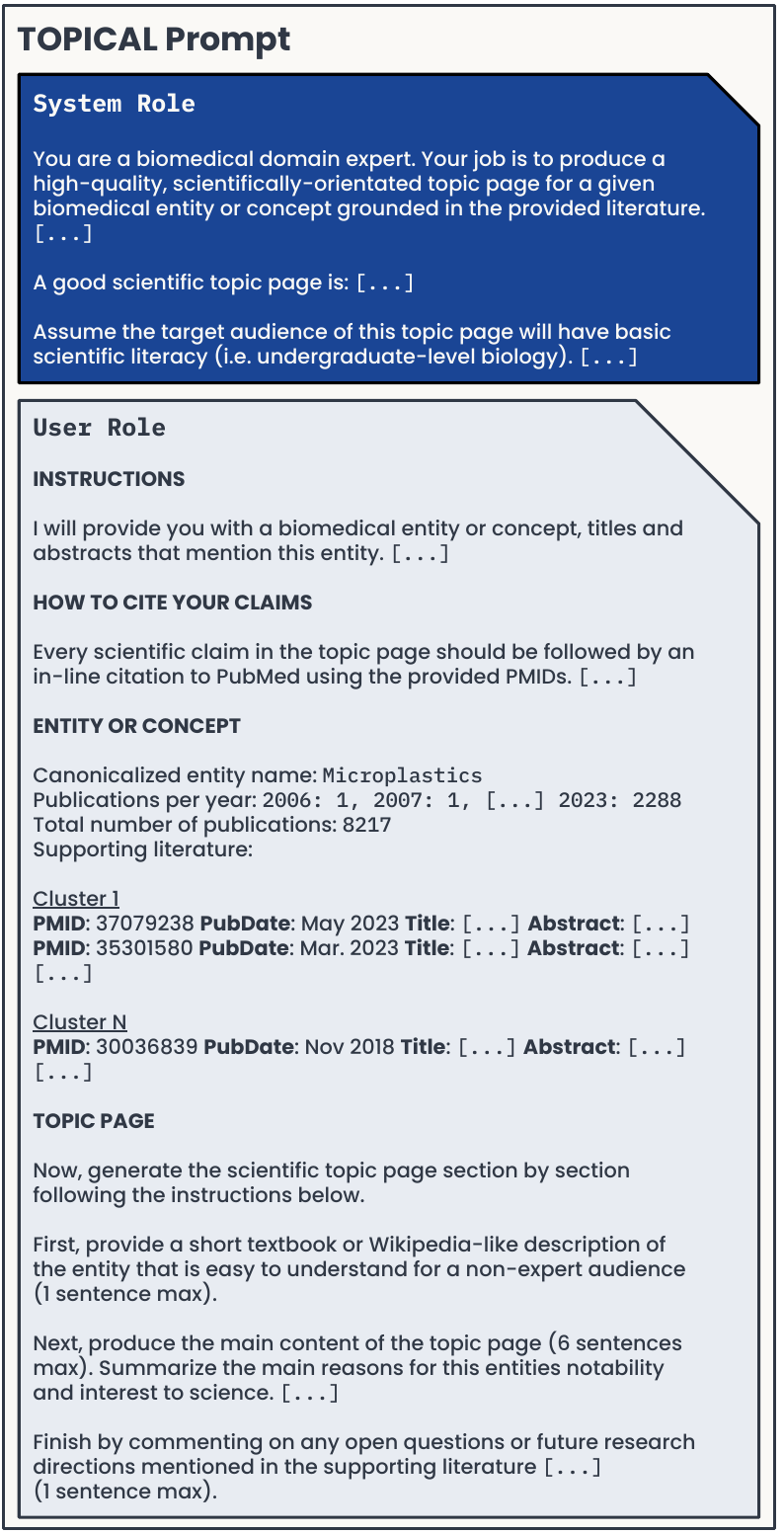}
\caption{Truncated example prompt. 
The prompt is divided into system and user roles. In the user role, we provide instructions about the input, how to cite a claim, details about the entity or concept like publication metadata, the sampled literature, and guidance about the expected sections and lengths for the topic page. Emphasis is provided for visualization purposes only.}
\label{fig:prompt}
\vspace{-1mm}
\end{figure}

We chose GPT-4\footnote{Specifically, the 06/13/2023 snapshot, ``\href{https://platform.openai.com/docs/models/gpt-4}{\texttt{gpt-4-0613}}''} as the LLM due to its state-of-the-art performance across many text generation tasks \citep{OpenAI2023GPT4TR}. We designed a prompt including natural language instructions, publication metadata, and the sampled titles and abstracts. The prompt is broken into system and user roles (truncated example in \autoref{fig:prompt}). In the system role, we provide instructions about the task and what constitutes a good topic page. The user role provides instruction about what the model will receive as input, followed by a description of how to cite its sources. We then provide information about the entity or concept, including the publications per year, total number of publications, and sampled titles and abstracts. These include a PMID and publication date and are sorted by decreasing cluster size. Finally, we provide instructions about the expected format of the topic page.

\begin{figure*}[t]
\includegraphics[width=\textwidth]{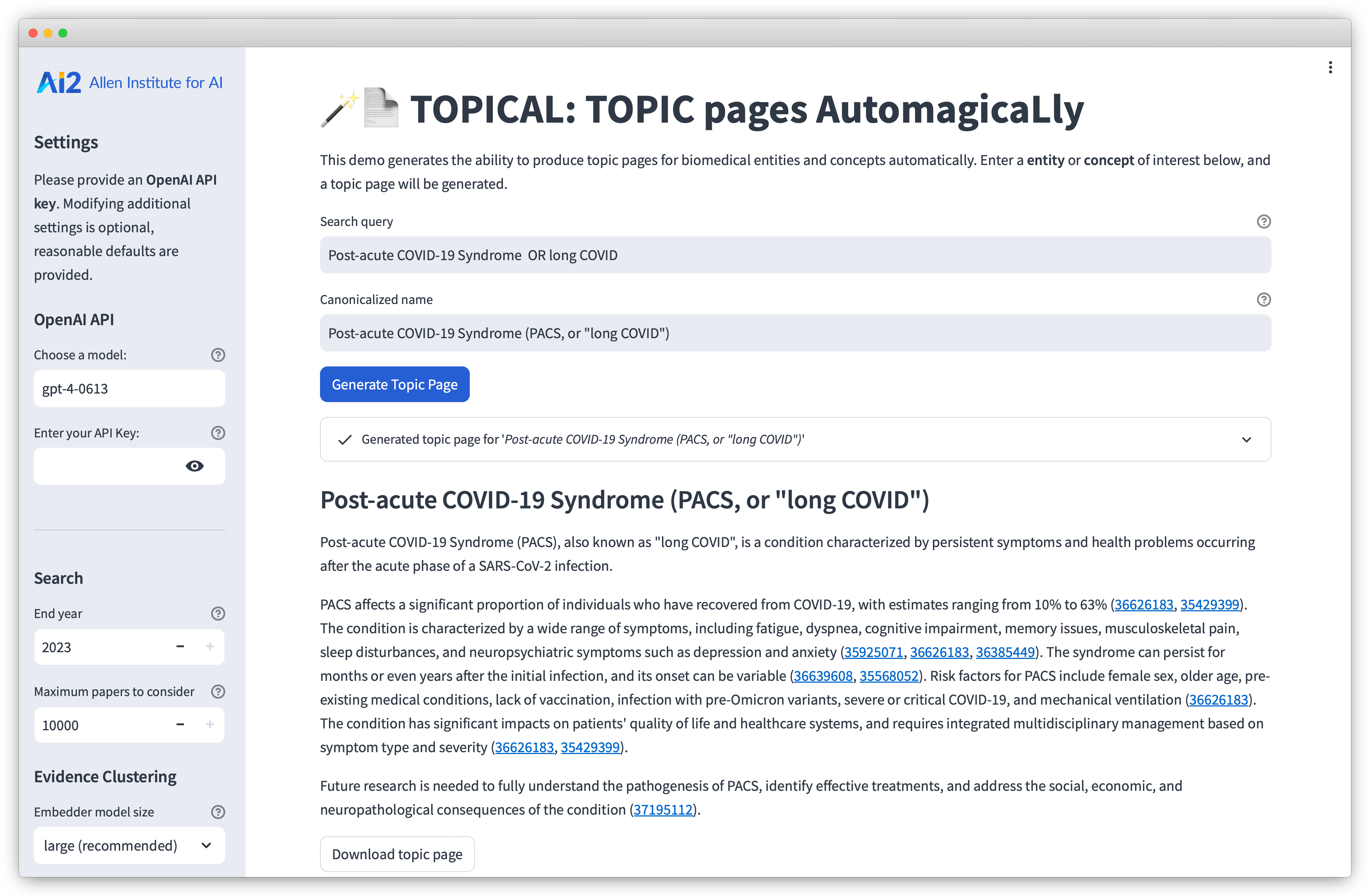}
\caption{TOPICAL web app. Given a search query for a biomedical entity or concept of interest and a canonicalized name, it automatically generates a topic page for the concept. An expandable section provides additional information, like a histogram of publication dates for the query and the number of clusters identified.}
\label{fig:web-app}
\vspace{-1mm}
\end{figure*}

\begin{figure}[t]
\includegraphics[width=\columnwidth]{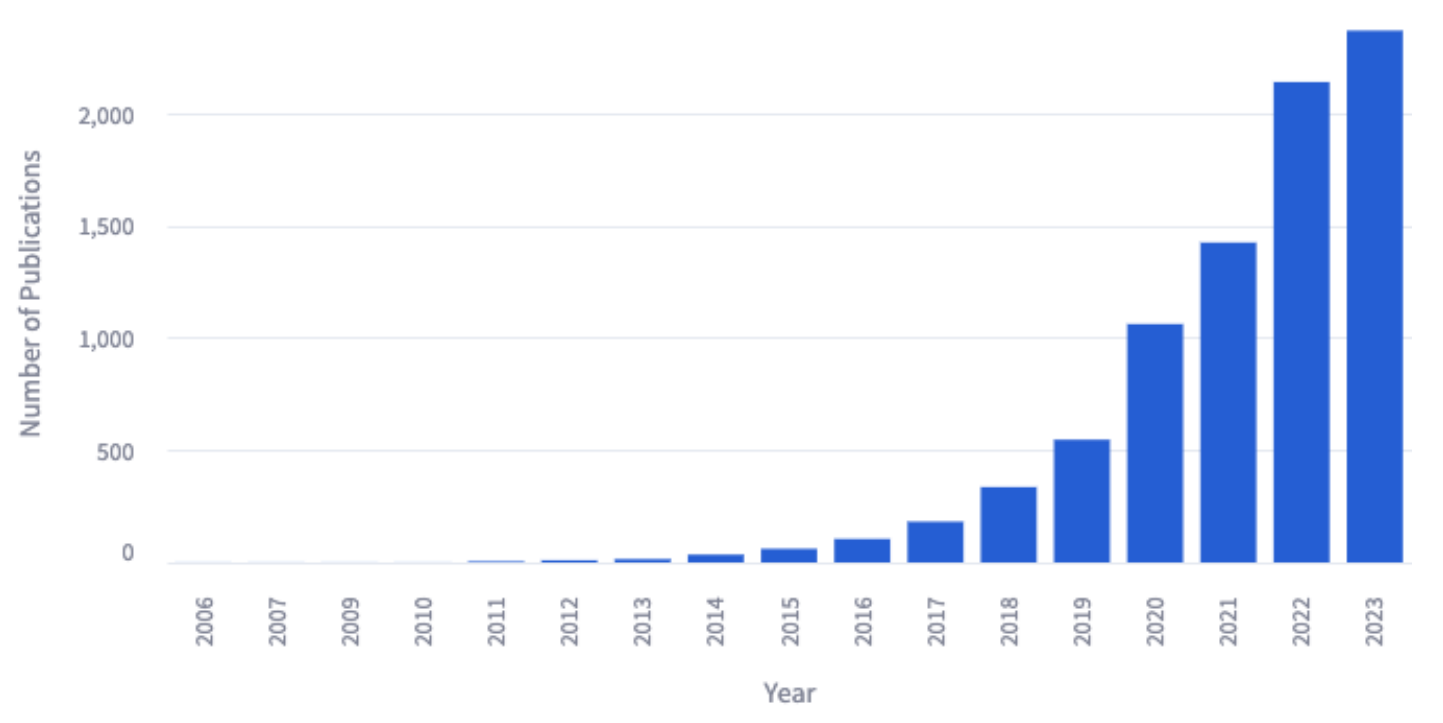}
\caption{Example publications per year histogram displayed to users for the entity: ``\textit{Microplastics}''.}
\label{fig:histogram}
\vspace{-2mm}
\end{figure}

The model is instructed to produce three sections: a \textbf{definition statement} (1-2 sentences), \textbf{main content} (5-8 sentences) and a concluding remark about \textbf{future research directions and open questions} (1 sentence). We model the components of our target topic pages based on the structure of existing scientific topic pages and the information researchers are likely to seek from a topical review. Curated topic pages typically begin with definitions,\footnote{e.g., \url{https://www.sciencedirect.com/topics}} so we also begin by generating a definition statement. Per the PRISMA guidelines for systematic reviews \citep{Page2020TheP2}, a primary goal of reviews is to provide ``syntheses of the state of knowledge in a field, from which future research priorities can be identified''; from this goal, we derive the main content, which summarizes the main directions of research, and future research directions.

We set \texttt{temperature} to 0.0, \texttt{max\_tokens} to 512, (the maximum tokens to generate for the topic page), and kept all other hyperparmeters of the OpenAI API at default values.\footnote{\url{https://platform.openai.com/docs/api-reference/completions}} The model's maximum context size is 8,192 tokens, which is approximately enough for the prompt instructions and 16 abstracts. To fit more abstracts into the prompt, we take only the first three and last two sentences of each, joining them with a ``\texttt{[TRUNCATE]}'' token. These sentences tend to be rich in  the type of content expected in a topic page, e.g., definition-like content, conclusions, major findings, and future directions.

\section{TOPICAL Web App} \label{web-app}

TOPICAL is available as a web app (see \autoref{fig:web-app} for an overview). The web app can be run locally as a standalone python package but is also publicly available at \url{https://s2-topical.apps.allenai.org}. A user first inputs a PubMed search query, which supports the full syntax of the PubMed Advanced Search Builder (see \autoref{appendix:advanced-search} for details). However, in most cases, simply inputting the entity or concept name directly and allowing ESearch to expand the query via automatic term mapping (ATM) works well, especially with respect to recall. A user can optionally provide alternative names for the entity, referred to as `canonicalized names,' which are provided to the LLM as additional context. Once a user clicks the button to generate a topic page, the search, embedding, clustering, and generation steps are executed. An expandable section in the app displays progress, as well as additional information about the search, e.g., any query expansions made via ATM and a histogram of publications per year (see \autoref{fig:histogram}). The generated topic pages can be downloaded as JSON files. A video demonstration of the system is available here: \url{https://youtu.be/hgnG7BnIeAY}.

\section{Human evaluation} \label{human-eval}

We conduct human evaluation to determine the quality of the automatically generated topic pages. The evaluation consists of two tasks, described below. All annotations were performed by three full-time, paid annotation specialists with undergraduate training spanning the biosciences, materials science, environmental science, and data science.

\subsection{Annotation Task 1: Topic Page}

In task 1, the goal was to evaluate the overall quality of the topic page along three facets, \textbf{relevance}, \textbf{accuracy}, and \textbf{coherence}, defined as:

\begin{itemize}[itemsep=0pt, topsep=2pt, leftmargin=10pt]
    \item \textbf{Relevance}: whether the topic page covers only important aspects of the entity or concept; unimportant or excess information is penalized
    \item \textbf{Accuracy}: whether the topic page is free of obvious factual errors or contradictory information
    \item \textbf{Coherence}: whether sentences and sections fit together and sound natural, with little to no redundancy within or across sections
\end{itemize}

\noindent We adapt these facets and their definitions from the summarization evaluation facets used in \citet{fabbri-etal-2021-summeval}; we assess Accuracy instead of Consistency due to the infeasibility of comparing a generated topic page against all input documents. Relevance and accuracy were assessed per topic page \textit{section} (definition statement, main content and future directions), while coherence was assessed globally. The annotation interface (see \autoref{appendix:annotation-interface}) displayed each section of the topic page, and the annotators were provided instructions about how to evaluate topic pages along each facet. The annotation interface also provided a link to the PubMed query issued when building the topic page. Annotators were instructed to follow the link and skim a handful of abstracts to familiarize themselves with the entity or concept before evaluation.

\begin{table*}[t]
\small
\centering
\caption{Results of human evaluation for annotation task 1. Total ratings for each facet and label are shown, along with agreement percentage. Two annotators rated 100 pages each (with 50\% overlap). Each facet for each section was rated on an ordinal scale: ``not'', ``somewhat'', or ``(yes)'' relevant/accurate/coherent.}
\label{tab:task-1}
\resizebox{\textwidth}{!}{%
\begin{tabular}{@{}lcccccccclc@{}}
\toprule
 & \multicolumn{2}{c}{Definition} &  & \multicolumn{2}{c}{Main content} &  & \multicolumn{2}{c}{Future directions} &  &  \\ \cmidrule(lr){2-3} \cmidrule(lr){5-6} \cmidrule(lr){8-9}
Rating          & relevant & accurate &  & relevant & accurate &  & relevant & accurate &                      & coherent \\ \midrule
missing/invalid & 0        & 0        &  & 0        & 0        &  & 0        & 0        &                      & --       \\
not             & 0        & 0        &  & 1        & 0        &  & 3        & 0        &                      & 0        \\
somewhat        & 4        & 1        &  & 7        & 0        &  & 15       & 0        &                      & 15       \\
yes             & 196      & 199      &  & 192      & 200      &  & 182      & 200      &                      & 185      \\ \midrule
Percent agreement    & 94     & 98     &  & 94     & 100      &  & 88     & 100      & \multicolumn{1}{c}{} & 82     \\ \bottomrule
\end{tabular}%
}
\vspace{-1mm}
\end{table*}

\begin{table}[t]
\small
\centering
\caption{Results of human evaluation for annotation task 2. Total ratings per label are shown, along with agreement percentage. Each annotator rated 100 citations, with 50\% overlap between annotators.}
\label{tab:task-2}
\resizebox{\columnwidth}{!}{%
\begin{tabular}{@{}lc@{}}
\toprule
Rating                           & Number of Ratings \\ \midrule
Incorrect (invalid)              & 0           \\
Incorrect (topically irrelevant) & 2           \\
Incorrect (topically relevant)   & 32          \\
Correct                          & 166         \\ \midrule
Percent agreement                     & 88          \\ \bottomrule
\end{tabular}%
}
\vspace{-1mm}
\end{table}

For each facet, annotators selected from one of three options: `not' \{relevant, accurate, coherent\}, `somewhat' \{relevant, accurate, coherent\} or simply: \{relevant, accurate, coherent\}. We included a fourth option for relevance and accuracy: `missing/invalid', in case the LLM failed to generate a particular topic page section.

\subsection{Annotation Task 2: Citations}

In task 2, the goal was to evaluate the relevance and sufficiency of model-provided citations. One citation from each topic page was sampled at random. Annotators were shown the citation in context and the cited article's title and abstract. They were instructed to annotate the citation as:

\begin{itemize}[noitemsep, topsep=0pt, leftmargin=10pt]
    \item {\color{MyGreen} \textbf{Correct}}: citation is topically \textit{relevant} (i.e., the cited article is about the target entity or concept) and provides sufficient evidence for the corresponding claim(s) in the topic page.
    \item {\color{MyOrange} \textbf{Incorrect (topically relevant)}}: citation is topically relevant but does not provide sufficient evidence for the corresponding claim(s).
    \item {\color{MyRed} \textbf{Incorrect (topically irrelevant)}}: citation is topically \textit{irrelevant}.
    \item {\color{MyGrey} \textbf{Incorrect (invalid)}}: citation is not valid, e.g. the PMID does not exist or was truncated.
\end{itemize}

\subsection{Choosing topics for evaluation}

In order to choose a broad selection of topics for evaluation, we collected all terms added to the MeSH vocabulary in the last 10 years (01/01/2013--16/10/2023, inclusive). We only include terms with a maximum tree depth\footnote{MeSH terms are organized in a polyhierarchical ontology, where more specific terms exist deeper in the tree.} of at least 7 as we found terms with a tree depth less than this tended to be overly broad and non-specific, e.g. ``\textit{Metadata}'', ``\textit{Rural Nursing}'', ``\textit{Infant Health}'', and ``\textit{Missed Diagnosis}''. The end result is 981 biomedical terms or concepts spanning a wide range of semantic types, including diseases (e.g. ``\textit{Charles Bonnet Syndrome}''), drugs (e.g. ``\textit{Modafinil}''), proteins (e.g. ``\textit{beta-Arrestin 1}), organisms (e.g ``\textit{Fallopia multiflora}''), cell types (e.g. ``\textit{Memory T Cells}'') and broader concepts like ``\textit{Glycemic Load}''. We sub-sampled from this set to produce the final list of entities for evaluation: 15 per annotator for the annotation pilots (with 100\% overlap) and 100 per annotator for the final evaluation (50\% overlap).

\section{Results} \label{results}

We find that the majority of topic pages are rated by our annotators as relevant, accurate, and coherent (\autoref{tab:task-1}), with high inter-annotator agreement (\(\ge82\%\)).\footnote{We report inter-annotator agreement as the percent agreement: (fraction of cases where annotators agree) / (total number of annotations).} We note that in no case did the model fail to output a topic page with the expected three-section structure. All sections received nearly perfect ratings for accuracy. The future direction section received the lowest rating for relevancy (18/200 ratings of `not' or `somewhat' relevant). Examining these instances reveals that the LLM often states vague or even obvious future directions, such as: ``\textit{[...] Future research is needed to further clarify the most effective use of this drug combination in the treatment of respiratory diseases}'' or ``\textit{Future research directions include further investigation into the exact mechanisms of resveratrol's action in diseases such as cancer and diabetes [...]}.'' We believe this reflects the inherent difficulty of identifying future research directions and open questions about a given topic. Coherence was the next lowest-rated aspect, with 15/200 ratings of `somewhat' coherent. The most common reason for this according to the annotators, by far, was extensive use of highly-specific jargon, making the topic page difficult to read as a non-expert.

Similarly, most model-provided citations were rated as correct (\autoref{tab:task-2}) with high inter-annotator agreement (\(\ge 88\%\)); in no case were the citations invalid, e.g., a hallucinated PMID. Most incorrect citations were marked as `Incorrect (topically relevant)' (32/200), denoting cases where the citation was \textit{on-topic}, but the cited article did not provide sufficient evidence for the corresponding claim(s).

\section{Conclusion}

In this paper, we present TOPICAL, a new approach for the automatic generation of high-quality scientific topic pages that leverages large language models (LLMs) and retrieval-augmented generation (RAG). We conducted an extensive human evaluation of 150 diverse topics from the biomedical literature and our annotators rated the vast majority of generated topic pages as relevant, accurate, and coherent; and model-provided citations as correct. Promising future directions include allowing users to provide custom instructions with respect to structure, focus and length of the automatically generated topic pages, and the investigation of open-source LLMs in place of the closed-source LLM we experimented with (GPT-4). We release a publicly available web app so that others can experiment with generating topic pages for entities or concepts of interest on demand.

\section*{Limitations}

\paragraph{Context window}

Due to the limited context window of GPT-4 (8192 tokens), our system only ingests a small fraction of literature for most entities or concepts. We tried to partially alleviate this through our clustering and sampling procedure, which is designed to encourage diversity in the selected literature while maintaining the representation of common research threads. A promising future direction is to explore the use of language models with significantly larger context windows, such as the recently announced GPT-4-turbo (128,000 tokens).

\paragraph{Unclear provenance}

Our evaluation is not able to determine to what degree the information in the resulting topic pages is derived from the learned weights of the language model itself, versus the retrieved literature. This is partially alleviated by requiring the language model to provide citations for all scientific claims, allowing a user to verify the information. 

\paragraph{Unit of retrieval}

We do not explore retrieving information other than titles or abstracts. It is possible that retrieving information on another level of granularity, e.g. sentences or ``chunks'', could improve the quality of the topic pages. It is also possible that extending retrieval to the full-content of a scientific paper could further improve quality. Determining the most performant granularity for the retrieval step is an exciting future direction.

\section*{Acknowledgements}

This research was enabled in part by support provided by the Digital Research Alliance of Canada (\href{https://alliancecan.ca/}{alliancecan.ca}) and Compute Ontario (\href{https://www.computeontario.ca/}{www.computeontario.ca}). We thank Jenna Sparks, Erin Bransom, and Bailey Kuehl for helping to conduct human evaluations. We thank the AI2 Reviz team for assistance with hosting the web application. We thank all internal and external reviewers for their thoughtful feedback, which improved earlier drafts of this manuscript.

\section*{Author Contributions}
\label{sec:contrib}

John Giorgi conducted data processing, engineered prompts, ran experiments, and implemented the evaluation. John also contributed to project scoping and ideation and wrote the paper with feedback from others. Sergey, Doug, and Lucy were project mentors, contributing equally to project scoping and experimental design and providing core ideas and direction throughout the course of the project and paper writing. Aman made technical contributions around scaling and hosting of the demo.

\bibliography{anthology,custom}


\appendix

\section{PubMed Advanced Search Builder}
\label{appendix:advanced-search}

TOPICAL supports the full syntax of the PubMed Advanced Search Builder. For example, to search for mentions of an entity in the title only: \\ [-3mm]

\noindent \texttt{Post-acute COVID-19 Syndrome[Title]} \\ [-3mm]

\noindent or for papers with the corresponding MeSH term: \\ [-3mm]

\noindent \texttt{Post-acute COVID-19 Syndrome[MeSH Terms]} \\ [-3mm]

\noindent Search terms can be further combined with \texttt{AND}, \texttt{OR} and \texttt{NOT} operators: \\ [-3mm]

\noindent  \texttt{Post-acute COVID-19 Syndrome[Title] AND Post-acute COVID-19 Syndrome[MeSH Terms]} \\ [-3mm]

\noindent However, in most cases, we found that simply inputting the entity or concept name directly and allowing ESearch to expand the query via automatic term mapping (ATM) works best, especially with respect to recall.

\begin{figure*}[t!]
\includegraphics[width=\textwidth]{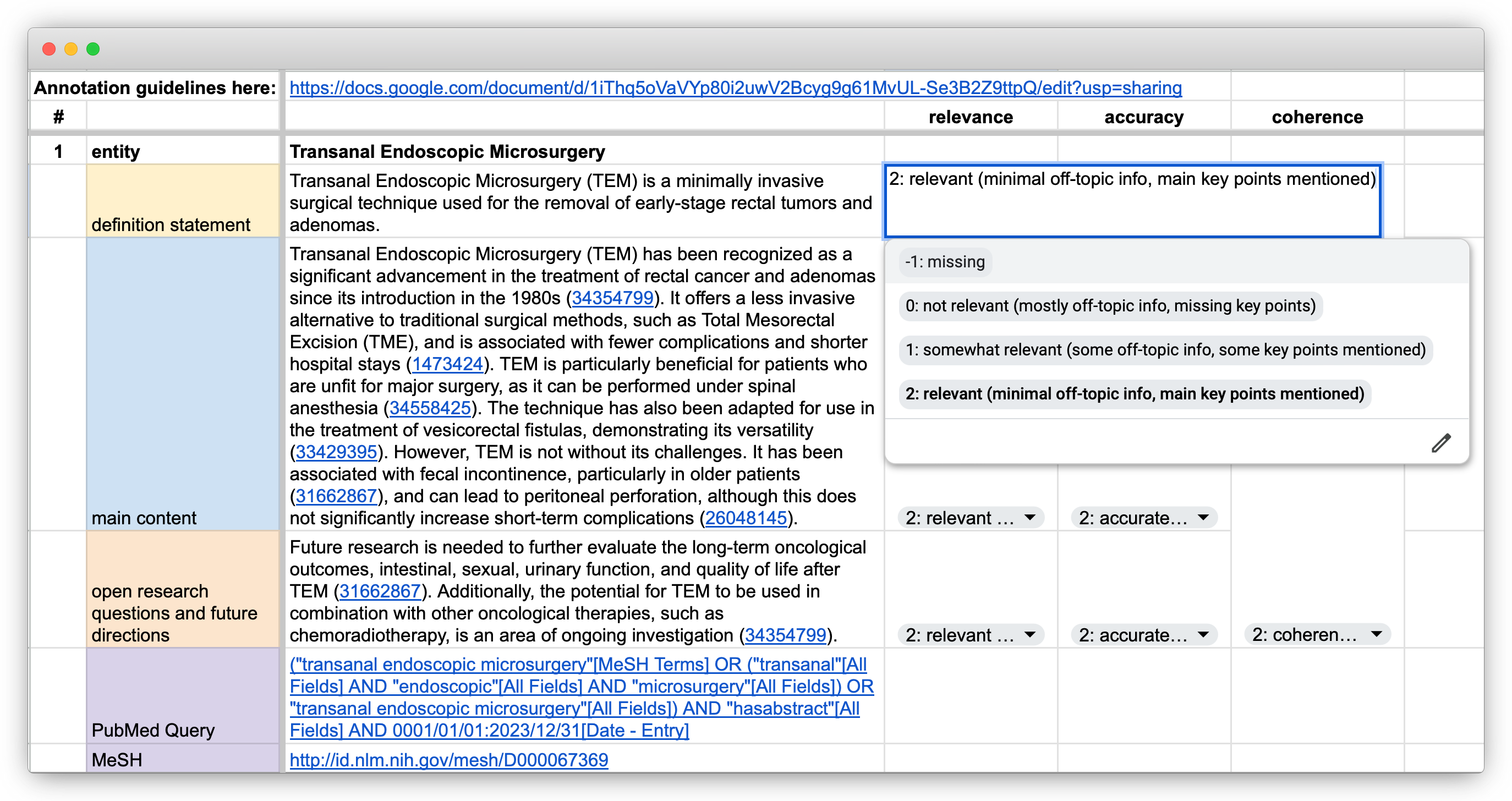}
\caption{Annotation interface for annotation task 1.}
\label{fig:task-1}
\end{figure*}

\section{Annotation Interface}
\label{appendix:annotation-interface}

In \autoref{fig:task-1}, we provide a screenshot of the annotation interface built in Google Sheets used for the human evaluation. Annotators were provided the contents of the topic page segmented into the three sections (definition statement, main content, and open research questions and future directions)

\section{Annotation pilots} \label{appendix:annotation-pilots}

Before the full evaluation, we ran 2 pilots with 3 annotators. The annotators evaluated the same 10 topic pages in the first pilot. We used their feedback to improve the annotation guidelines and identify the main sources of inter-annotator disagreement. Most notably, task 2 originally had annotators identify all unique claims in each section of the topic page and then annotate each following the guidelines. This turned out to be overly time-intensive, and determining the specific number of claims had a very low-inter-annotator agreement. Task 2 was therefore simplified by randomly sampling one citation in the topic page and having the annotators assess its relevance and sufficiency. We then ran a second pilot on a new set of 5 topic pages to finalize the annotation guidelines and identify any remaining sources of significant annotator disagreement.

\end{document}